\pdfoutput=1

\documentclass[11pt]{article}

\usepackage{ACL2023} 

\newcommand{\citeg}[1]{\citeauthor{#1}'s \citeyearpar{#1}} 

\usepackage{times}
\usepackage{latexsym}

\usepackage[T1]{fontenc}

\usepackage[utf8]{inputenc}

\usepackage{microtype}

\usepackage{inconsolata}

\usepackage{caption}
\usepackage{subcaption}

\usepackage{graphicx}

\usepackage{arydshln}
\usepackage{multirow}

\usepackage{booktabs}

\usepackage{amsmath}
\usepackage{float}

\title{Evaluating the Effectiveness of Natural Language Inference for \\ Hate Speech Detection in Languages with Limited Labeled Data}

\author{Janis Goldzycher \hspace{0.8cm} Moritz Preisig \hspace{0.8cm} Chantal Amrhein \hspace{0.8cm} Gerold Schneider\\
  Department of Computational Linguistics \\
  University of Zurich  \\
  \texttt{\{goldzycher,amrhein,gschneid\}@cl.uzh.ch, moritz.preisig@uzh.ch} \\
}

\begin{document}
\maketitle
\begin{abstract}
Most research on hate speech detection has focused on English where a sizeable amount of labeled training data is available. 
However, to expand hate speech detection into more languages, approaches that require minimal training data are needed. 
In this paper, we test whether natural language inference (NLI) models which perform well in zero- and few-shot settings can benefit hate speech detection performance in scenarios where only a limited amount of labeled data is available in the target language. 
Our evaluation on five languages demonstrates large performance improvements of NLI fine-tuning over direct fine-tuning in the target language. However, the effectiveness of previous work that proposed intermediate fine-tuning on English data is hard to match. Only in settings where the English training data does not match the test domain, can our customised NLI-formulation outperform intermediate fine-tuning on English.
Based on our extensive experiments, we propose a set of recommendations for hate speech detection in languages where minimal labeled training data is available.
\footnote{We make our code publically available at \url{https://github.com/jagol/xnli4xhsd}.} 
\end{abstract}

\begin{figure}[t]
\center
\includegraphics[width=0.9\linewidth]{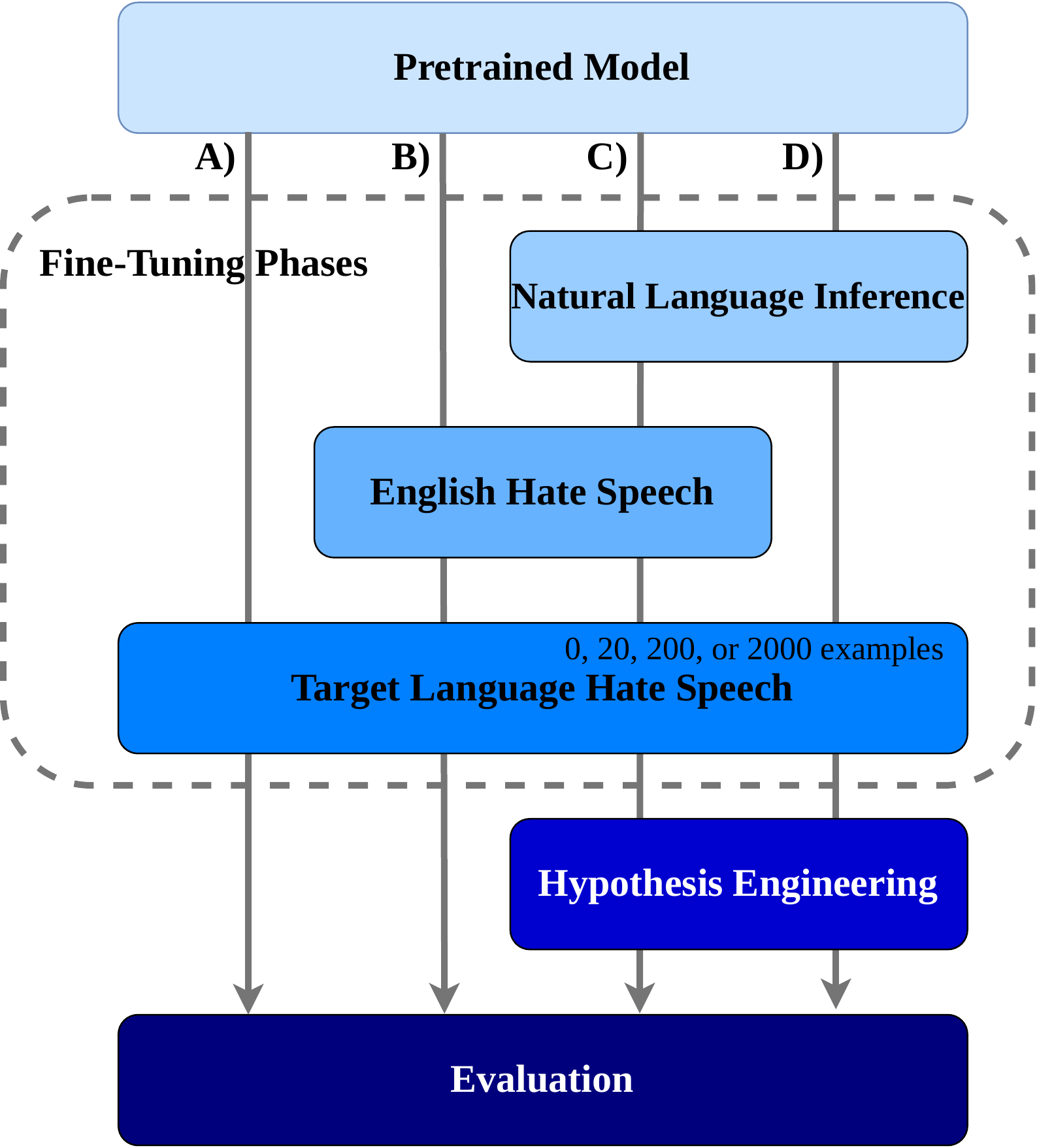}
\caption{Approaches evaluated in this paper: A) Standard fine-tuning. B) Intermediate fine-tuning on English hate speech as proposed by \citet{rottger-etal-2022-data}. Via C) and D), we explore natural language inference as an additional intermediate fine-tuning step. The natural language inference formulation further allows hypothesis engineering \cite{goldzycher-schneider-2022-hypothesis}.} 
\label{fig:approaches}
\end{figure}

\section{Introduction}

Hate speech is a global issue that transcends linguistic boundaries, but the majority of available datasets for hate speech detection are in English \cite{poletto_resources_2021, yin_towards_2021}. This limits the capabilities of automatic content moderation and leaves most language communities around the world underserved. Creating labeled datasets is not only slow and expensive but also risks psychological impacts on the annotators \cite{kirk-etal-2022-handling}. Although the number of non-English datasets is increasing, most languages still have limited or no datasets available \cite{poletto_resources_2021}. Consequently, there is a pressing need to develop methods that can efficiently expand hate speech detection into languages with less labeled data.
Repurposing natural language inference models for text classification leads to well-performing zero-shot and few-shot classifiers \cite{yin_benchmarking_2019}. 
Recently, \citet{goldzycher-schneider-2022-hypothesis} showed that zero-shot NLI-based setups can outperform standard few-shot finetuning for English hate speech detection. This raises the question of whether natural language inference can be used to expand hate speech detection into more languages in a data-efficient manner.

In this paper, we aim to systematically evaluate the effectiveness of NLI fine-tuning for languages beyond English where we do not have an abundance of labeled training data. 
We give an overview of the experiment setup and compared approaches in Figure \ref{fig:approaches}. 
Unlike \citet{goldzycher-schneider-2022-hypothesis}, and inspired by \citet{rottger-etal-2022-data}, we do not restrict ourselves to a zero-shot setup but further analyse the usefulness of an NLI-formulation when we have access to a limited amount of labeled examples in the target language as well as additional English data for intermediate fine-tuning. We believe that this mirrors a more realistic setup and allows us to offer clear recommendations for best practices for hate speech detection in languages with limited labeled data. 

Our experiments with $0$ up to $2000$ labeled examples across five target languages (Arabic, Hindi, Italian, Portuguese, and Spanish) demonstrate clear benefits of an NLI-based formulation in zero-shot and few-shot settings compared to standard few-shot fine-tuning in the target language. While \citeg{rottger-etal-2022-data} approach of fine-tuning on English data before standard few-shot learning on the target language proves to be a strong baseline, we reach similar performance when fine-tuning NLI-based models on intermediate English data. Building on the results by \citet{goldzycher-schneider-2022-hypothesis}, who showed that targeted hypothesis engineering can help avoid common classification errors with NLI-based models, we find that such strategies offer an advantage in scenarios where we have expert knowledge about the domain but no in-domain English data for intermediate fine-tuning.

Overall, our contributions are the following:
\begin{enumerate}
\itemsep0em 
    \item We are able to reproduce the results of \citet{rottger-etal-2022-data}, demonstrating the validity of their approach.
    \item We evaluate NLI fine-tuning for expanding hate speech detection into more languages and find it to be beneficial if no English labeled data is available for intermediate fine-tuning.
    \item We evaluate NLI-based models paired with hypothesis engineering and show that we can outperform previous work in settings where we have knowledge about the target domain but no domain-specific labeled English data.
\end{enumerate}

\section{Related Work}

Hate speech is commonly defined as attacking, abusive or discriminatory language that targets protected groups or an individual for being a member of a protected group. A protected group is defined by characteristics such as gender, sexual orientation, disability, race, religion, national origin or similar \cite{fortuna_survey_2018, poletto_resources_2021, vidgen-etal-2021-learning, yin_towards_2021}.
The automatic detection of hate speech is typically formulated as a binary text classification task with short texts, usually social media posts and comments, as input \cite{founta_large_2018}. Despite most work being focused on English \cite{founta_large_2018}, in the last years there has been a growing trend to expand into more languages \cite{mandl_overview_2019, mandl_overview_2020, rottger-etal-2022-multilingual, yadav_lahm_2023}.

In what follows, we first review the relevant literature for multi- and cross-lingual hate speech detection, and then move on to the previous work in zero-shot and few-shot text classification with a specific focus on NLI-based methods, and finally focus on hypothesis engineering \cite{goldzycher-schneider-2022-hypothesis}. 

\subsection{Multi- and Cross-lingual Hate Speech Detection}
The scarcity of labeled datasets for hate speech detection in non-English languages has led to multiple approaches addressing this problem using meta-learning \cite{mozafari_cross-lingual_2022}, active learning \cite{kirk-etal-2022-data}, label-bootstrapping \cite{bigoulaeva_label_2023}, pseudo-label fine-tuning \cite{zia_improving_2022}, and multi-task learning using multilingual auxiliary tasks such as dependency parsing, named entity recognition and sentiment analysis \cite{montariol-etal-2022-multilingual}. 

The most important research for our investigation is the study conducted by \citet{rottger-etal-2022-data}. In their work, the authors train and evaluate models across five distinct languages: Arabic, Hindi, Italian, Portuguese, and Spanish. Their findings reveal that by initially fine-tuning multilingual models on English hate speech and subsequently fine-tuning them with labeled data in the target language, they achieve significant performance improvements in low-resource settings compared to only fine-tuning a monolingual model in the target language. In this study, we adopt their evaluation setup, reproduce their results and compare our results directly to their approach. For this reason, we will elaborate on and reference the specifics of their experimental setup throughout Section \ref{sec:experiment-setup}.

\subsection{Zero-Shot and Few-Shot Classification}
The development of language models which serve as a foundation for fine-tuning rather than training from scratch, has facilitated the implementation of zero-shot and few-shot text classification approaches, such as prompting \cite{liu_pre-train_2021} and task descriptions \cite{raffel_exploring_2020}. These techniques transform the target task into a format similar to the pre-training task and are typically employed in conjunction with large language models. Following this scheme, \citet{chiu_detecting_2021} leverage GPT-3 to detect hate speech with the prompts ``Is this text racist?'' and ``Is this text sexist?''.

In contrast to prompting, NLI-based prediction refers to reformulating the target task into an NLI task and thus into a (previous) fine-tuning task. In this setup, a model receives a premise and a hypothesis and is tasked with predicting whether the premise entails the hypothesis, contradicts it, or is neutral towards it.
\citet{yin_benchmarking_2019} were the first to demonstrate the effectiveness of such an approach. They used an NLI model for zero-shot topic classification by inputting the text to be classified as the premise and constructing a hypothesis for each topic in the form of ``This text is about <topic>''. A prediction of \textit{entailment} is to be interpreted as the input text belonging to the topic in the given hypothesis.
\citet{wang_entailment_2021} demonstrated that this task reformulation benefits few-shot learning scenarios for various tasks, including offensive language identification.

\subsection{Hypothesis Engineering}
\label{subsec:hypothesis-engineering}
\citet{goldzycher-schneider-2022-hypothesis} pair an input text with multiple hypotheses in order to let an NLI model predict different aspects of the input text. They then use a rule-based approach to combine these predicted aspects into a final prediction for the hate speech label.
More specifically, they distinguish between a main hypothesis and auxiliary hypotheses. The main hypothesis claims that the input text contains hate speech. The auxiliary hypotheses claim various relevant aspects such as that the input text is about a protected group in order to correct mispredictions of the main hypothesis.
To find effective hypothesis combinations they conduct an error analysis on the English HateCheck \cite{rottger-etal-2021-hatecheck} dataset and propose four \textit{strategies} based on this error analysis: 
\begin{description}
\itemsep0em 
    \item[Filtering by target:] Avoid false positives by predicting if any protected group is targeted in the input text.
    \item[Filtering reclaimed slurs:] Avoid false positives by predicting indicators that a slur is used in a reclaimed fashion. Indicators used are: the speaker talks about themself or positive sentiment. 
    \item[Filtering counterspeech:] Avoid false positives by recognizing when another speech act is referenced, predicting if that speech act is hate speech and predicting the stance towards the referenced speech. 
    \item[Catching dehumanizing comparisons:] Avoid false negatives by checking if a protected group and negatively associcated animals appear together in a sentence with a negative sentiment. 
\end{description}

In our experiments, we will evaluate the effectiveness of the first three strategies for NLI-based hate speech detection. However, we will exclude the fourth strategy due to its lack of clear benefits in \citet{goldzycher-schneider-2022-hypothesis}. More implementation details are provided in Section \ref{subsec:hypo-eng} and Appendix \ref{appsec:hypo-eng-details}.

\begin{figure*}[t]
\center
\includegraphics[width=\linewidth]{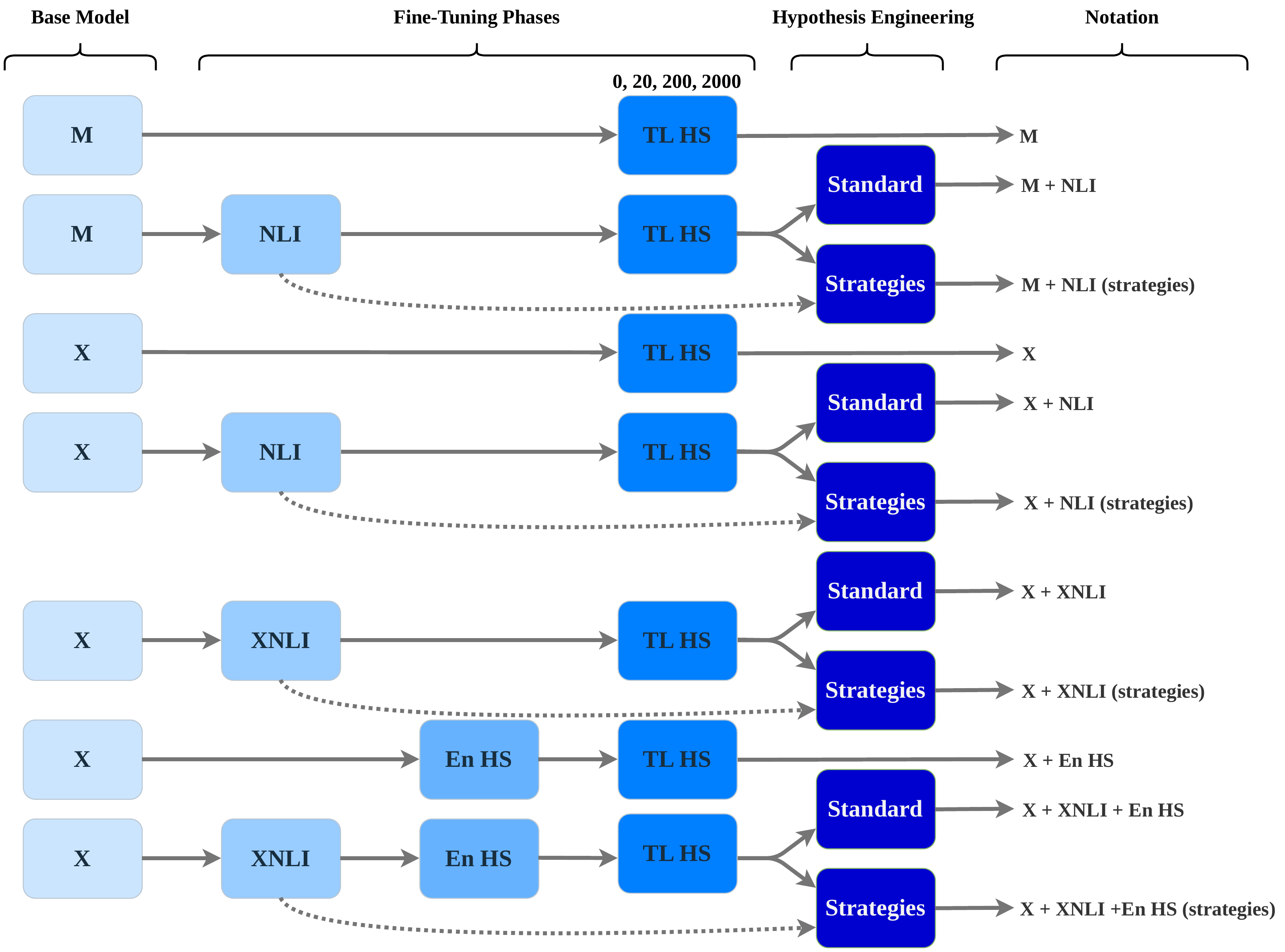}
\caption{Experiment Setup. The base model is either monolingual \textbf{M} or multilingual \textbf{X}. In a first optional training phase, the model is fine-tuned either on the full XNLI dataset (\textbf{XNLI}) or on the subset of XNLI that is in the target language (\textbf{NLI}). A second optional training phase follows, in which a model is fine-tuned on an English hate speech dataset (\textbf{En HS}). Note that \textbf{En HS} is a stand-in for the specific English dataset the model is fine-tuned on, i.e. either \textbf{DEN}, if it is fine-tuned on \citet{vidgen-etal-2021-learning} or \textbf{FEN}, if it is fine-tuned on \citet{founta_large_2018}. Finally, each model is fine-tuned on 0, 20, 200, or 2000 examples of a hate speech dataset in the target language (\textbf{TL HS}). If the model was fine-tuned on NLI, we either evaluate in a standard fashion with one hypothesis (\textbf{Standard}), or with hypothesis engineering strategies (\textbf{Strategies}). The dotted arrows show that for auxiliary hypotheses of the strategies a model version that was only fine-tuned on NLI is used. Further explanations of the dataset and model notation are provided in Section \ref{sec:experiment-setup}.}
\label{fig:experiment-setup}
\end{figure*}

\begin{table*}[t]
\setlength\tabcolsep{4pt}
\resizebox{\textwidth}{!}{
\begin{tabular}{lrrrrrrrrrrrrrrrr}
\toprule
 & \multicolumn{3}{c}{BAS19\_ES} & \multicolumn{3}{c}{FOR19\_PT} & \multicolumn{3}{c}{HAS21\_HI} & \multicolumn{3}{c}{OUS19\_AR} & \multicolumn{3}{c}{SAN20\_IT} & \multicolumn{1}{l}{Avg. Diff.} \\
\multicolumn{1}{r}{N} & \multicolumn{1}{c}{20} & \multicolumn{1}{c}{200} & \multicolumn{1}{c}{2000} & \multicolumn{1}{c}{20} & \multicolumn{1}{c}{200} & \multicolumn{1}{c}{2000} & \multicolumn{1}{c}{20} & \multicolumn{1}{c}{200} & \multicolumn{1}{c}{2000} & \multicolumn{1}{c}{20} & \multicolumn{1}{c}{200} & \multicolumn{1}{c}{2000} & \multicolumn{1}{c}{20} & \multicolumn{1}{c}{200} & \multicolumn{1}{c}{2000} & \multicolumn{1}{r}{} \\ 
\midrule
M & 0.48 & 0.67 & \textbf{0.84} & 0.46 & 0.62 & 0.71 & 0.46 & 0.49 & 0.56 & 0.45 & 0.55 & 0.68 & 0.40 & 0.70 & \textbf{0.78} & \multicolumn{1}{r}{-0.03} \\
X & 0.40 & 0.61 & 0.82 & 0.45 & 0.52 & 0.71 & 0.46 & 0.46 & \textbf{0.59} & 0.45 & 0.55 & \textbf{0.69} & 0.40 & 0.66 & 0.76 & \multicolumn{1}{r}{-0.03} \\
X + DEN & \textbf{0.66} & \textbf{0.75} & 0.83 & \textbf{0.63} & 0.68 & 0.71 & 0.51 & 0.53 & 0.58 & 0.51 & 0.64 & 0.67 & \textbf{0.64} & \textbf{0.71} & 0.76 & \multicolumn{1}{r}{-0.01} \\
X + FEN & 0.54 & 0.70 & 0.82 & \textbf{0.63} & \textbf{0.69} & \textbf{0.72} & \textbf{0.54} & \textbf{0.55} & \textbf{0.59} & \textbf{0.59} & \textbf{0.66} & 0.68 & \textbf{0.64} & \textbf{0.71} & 0.75 & \multicolumn{1}{r}{-0.01} \\
X + KEN & 0.63 & 0.72 & 0.82 & \textbf{0.63} & 0.68 & 0.71 & 0.52 & \textbf{0.55} & \textbf{0.59} & \textbf{0.59} & 0.65 & \textbf{0.69} & \textbf{0.64} & \textbf{0.71} & 0.75 & 0.00 \\
\midrule
Avg. Diff. & -0.03 & -0.02 & 0.02 & -0.01 & -0.05 & -0.01 & -0.01 & -0.02 & -0.01 & -0.01 & -0.06 & -0.02 & 0.00 & -0.02 & -0.01 & \multicolumn{1}{r}{-0.02} \\
\bottomrule
\toprule
 & \multicolumn{3}{c}{HateCheck\_ES} & \multicolumn{3}{c}{HateCheck\_PT} & \multicolumn{3}{c}{HateCheck\_Hi} & \multicolumn{3}{c}{HateCheck\_Ar} & \multicolumn{3}{c}{HateCheck\_It} & \multicolumn{1}{c}{Avg. Diff.} \\
\multicolumn{1}{r}{N} & \multicolumn{1}{c}{20} & \multicolumn{1}{c}{200} & \multicolumn{1}{c}{2000} & \multicolumn{1}{c}{20} & \multicolumn{1}{c}{200} & \multicolumn{1}{c}{2000} & \multicolumn{1}{c}{20} & \multicolumn{1}{c}{200} & \multicolumn{1}{c}{2000} & \multicolumn{1}{c}{20} & \multicolumn{1}{c}{200} & \multicolumn{1}{c}{2000} & \multicolumn{1}{c}{20} & \multicolumn{1}{c}{200} & \multicolumn{1}{c}{2000} & \multicolumn{1}{l}{} \\
\midrule
M & 0.44 & 0.48 & 0.59 & 0.35 & 0.50 & 0.62 & 0.23 & 0.23 & 0.23 & 0.26 & 0.23 & 0.24 & 0.28 & 0.39 & 0.54 & 0.01 \\
X & 0.40 & 0.31 & 0.60 & 0.31 & 0.32 & 0.64 & 0.34 & 0.23 & 0.24 & 0.30 & 0.23 & 0.24 & 0.36 & 0.42 & 0.59 & 0.00 \\
X + DEN & \textbf{0.82} & \textbf{0.82} & \textbf{0.80} & \textbf{0.79} & \textbf{0.81} & \textbf{0.78} & \textbf{0.57} & \textbf{0.38} & \textbf{0.36} & \textbf{0.61} & \textbf{0.48} & \textbf{0.35} & \textbf{0.82} & \textbf{0.81} & \textbf{0.79} & 0.01 \\
X + FEN & 0.57 & 0.60 & 0.64 & 0.56 & 0.59 & 0.62 & 0.34 & 0.30 & 0.34 & 0.35 & 0.34 & 0.29 & 0.54 & 0.55 & 0.58 & -0.01 \\
X + KEN & 0.57 & 0.58 & 0.63 & 0.62 & 0.64 & 0.64 & 0.40 & 0.30 & 0.31 & 0.31 & 0.30 & 0.29 & 0.53 & 0.59 & 0.62 & 0.02 \\
\midrule
Avg. Diff. & 0.02 & 0.00 & -0.01 & 0.02 & -0.04 & 0.02 & 0.02 & -0.04 & 0.03 & -0.02 & 0.00 & 0.02 & 0.02 & -0.01 & 0.06 & 0.01 \\
\bottomrule
\end{tabular}
}
\caption{Reproduction of results on held-out test sets and the Multilingual HateCheck. ``Avg. Diff.'' contains the average difference to the original results by \citet{rottger-etal-2022-data} per row and column respectively.}
\label{tab:repro-results}
\end{table*}

\section{Experiment Setup}
\label{sec:experiment-setup}
Our experiment setup is largely based on the one created by \citet{rottger-etal-2022-data} and can be seen in Figure \ref{fig:experiment-setup}. In what follows, we first describe their setup, which we replicated as a baseline for our results. We then describe how we expanded their setup for the NLI-based experiments.

\begin{figure*}[t]
\center
\includegraphics[width=\linewidth]{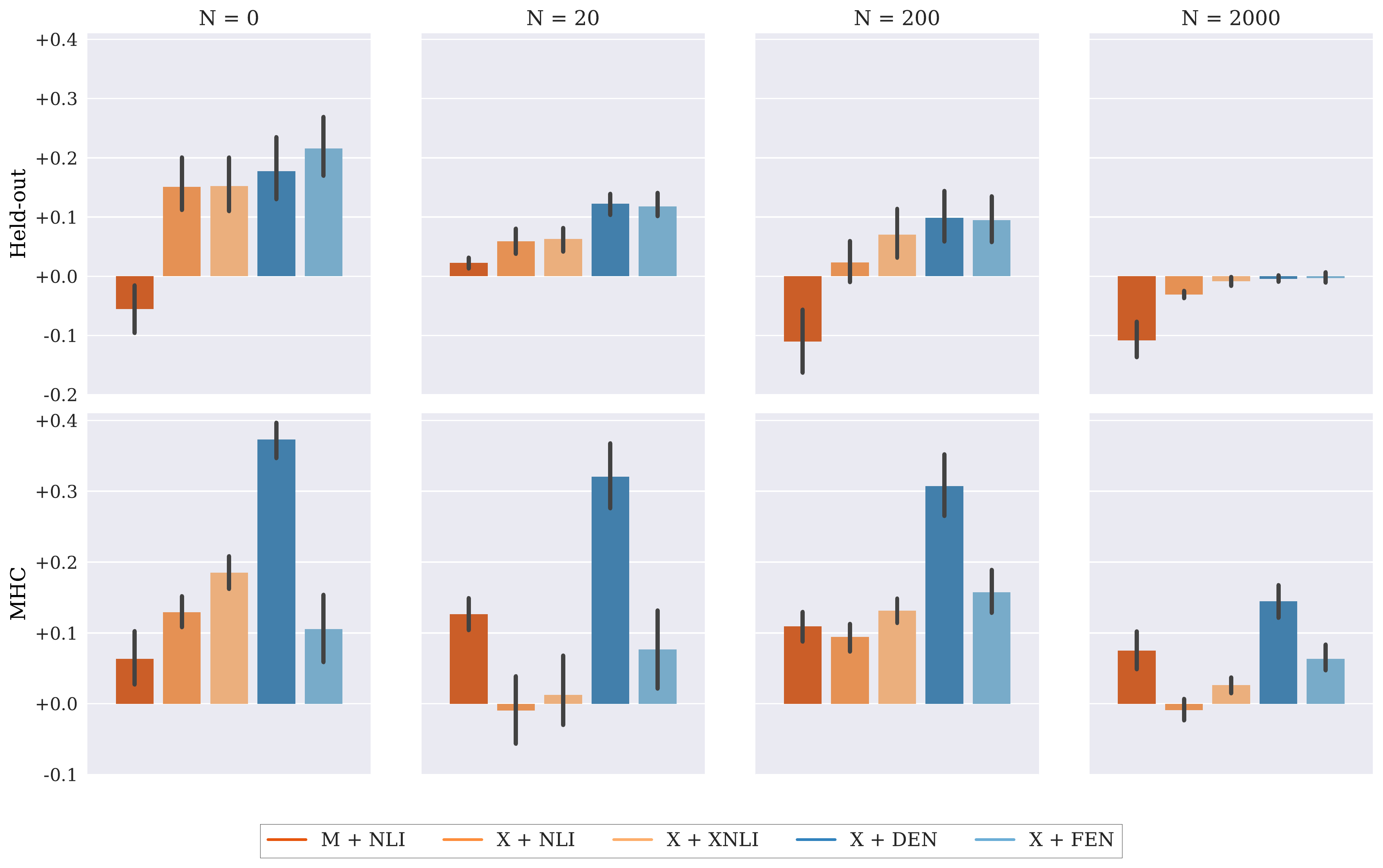}
\caption{Evaluation of NLI fine-tuning on languages that are in XNLI, namely, Arabic, Hindi, and Spanish. The figure shows the absolute difference in macro-$F_1$ score when adding an intermediate fine-tuning step to fine-tuning in the target language (i.e. the difference to \textbf{M} and \textbf{X}, respectively).}
\label{fig:baseline-nli-english-hs-seen-langauges}
\end{figure*}

\subsection{Reproducing \citet{rottger-etal-2022-data}}
\paragraph{Data} The authors use three English hate speech datasets \cite[abbreviated as \textbf{FEN}, \textbf{KEN}, and \textbf{DEN}, respectively]{founta_large_2018, kennedy-etal-2020-contextualizing, vidgen-etal-2021-learning} which are all downsampled to 20,000 examples. \textbf{FEN} and \textbf{KEN} are sourced from Twitter and \textbf{DEN} consists of human-created adversarial examples. They further make use of five Twitter datasets in the respective target languages: \citet{basile-etal-2019-semeval} 
for Spanish (\textbf{BAS19\_ES}), \citet{fortuna-etal-2019-hierarchically} for Portuguese (\textbf{FOR19\_PT}), \citet{10.1145/3503162.3503176} for Hindi (\textbf{HAS21\_HI}), \citet{ousidhoum-etal-2019-multilingual} for Arabic (\textbf{OUS19\_AR}), and \citet{manuela2020haspeede} for Italian (\textbf{SAN20\_IT}).
The Multilingual HateCheck (MHC) \cite{rottger-etal-2022-multilingual} is used for additional, complementary evaluation.  This suite of synthetic, evaluation-only datasets, that covers a range of typical, but hard to classify, cases for hate speech detection is available in all five target languages. 
The datasets are further described in Appendix \ref{appsec:datasets}.

\paragraph{Preprocessing} All datasets are cleaned with the same preprocessing steps that have been used for training of XLM-T \citet{barbieri-etal-2022-xlm}. These consist of replacing all URLs with ``https'' and all usernames (strings starting with an ``@'') with ``@user''. 
Further, the authors downsampled the non-hate speech class in \textbf{FEN} and \textbf{KEN} such that the relative frequency of hate speech increased from 5.0\% to 22\% and from 29.3\% to 50\% respectively. 

\paragraph{Models}
\citet{rottger-etal-2022-data} use \href{https://huggingface.co/cardiffnlp/twitter-xlm-roberta-base}{Twitter-XLM-RoBERTa-base} \cite{barbieri-etal-2022-xlm}, typically abbreviated to XLM-T, as a multi-lingual base model. This model is derived from \href{https://huggingface.co/xlm-roberta-base}{XLM-R} \cite{conneau-etal-2020-unsupervised} and has been further pre-trained on a multilingual Twitter corpus.
Further they use the following mono-lingual base-models: \href{https://huggingface.co/aubmindlab/bert-base-arabertv02}{AraBERT-v2} \cite{antoun2020arabert} for Arabic, \href{https://huggingface.co/neuralspace-reverie/indic-transformers-hi-bert}{Hindi BERT} for Hindi, \href{https://huggingface.co/Musixmatch/umberto-commoncrawl-cased-v1}{UmBERTo} \cite{musixmatch-2020-umberto} for Italian, \href{https://huggingface.co/neuralmind/bert-base-portuguese-cased}{BERTimbau} \cite{souza2020bertimbau} for Portuguese, and \href{https://huggingface.co/pysentimiento/robertuito-base-cased}{RoBERTuito} \cite{perez-etal-2022-robertuito} for Spanish. 

\paragraph{Training}

The training procedure consists of two fine-tuning phases: In the optional first phase, specifically proposed by \citet{rottger-etal-2022-data}, a model is fine-tuned on English hate speech (\textbf{En HS}). In the second phase, the model is fine-tuned on $N$ target language hate speech examples (\textbf{TL HS}), where $N \in \{10, 20, 30, 40, 50, 100, 200, 300, 400, 500, $ $1000, 2000\}$.\footnote{As explained later in Section \ref{sec:evaluation}, we only train and evaluate at $N \in \{0, 20, 200, 2000\}$.}
The training setup and the corresponding model notation are included in Figure \ref{fig:experiment-setup} as \textbf{X + En HS}. Note that \textbf{En HS} is a stand-in for the specific English dataset the model is fine-tuned on, i.e. either \textbf{FEN}, \textbf{KEN}, or \textbf{DEN}.
Additionally, \citet{rottger-etal-2022-data} compare against the baselines of fine-tuning a monolingual model directly on the target language (\textbf{M}) and fine-tuning a multilingual model directly on the target language (\textbf{X}).
Training specifics, including hyperparameters, are provided in Appendix \ref{appsec:training-details}.
 
\begin{figure*}[t]
\center
\includegraphics[width=\linewidth]{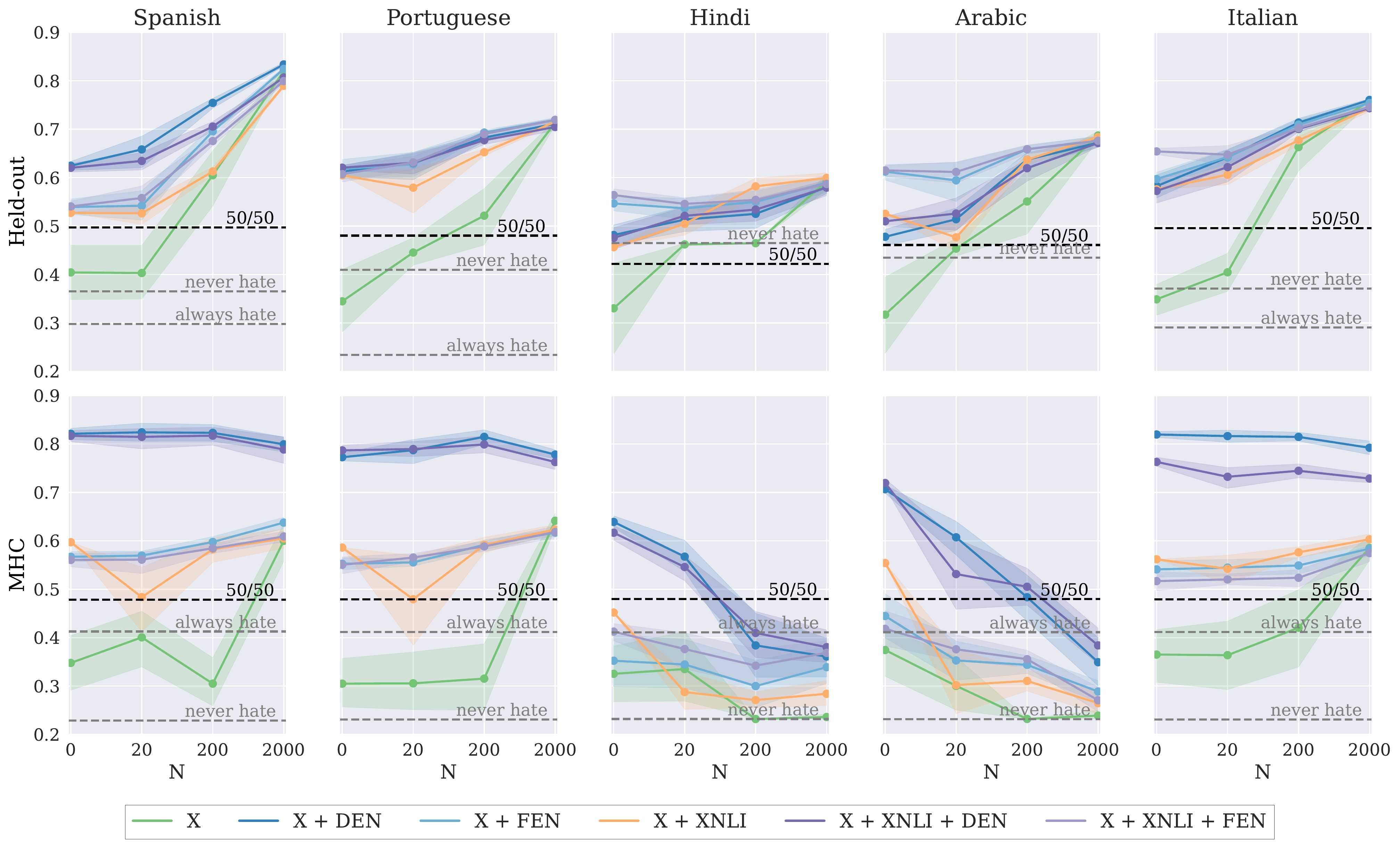}
\caption{Evaluation of NLI fine-tuning on top of English hate speech fine-tuning. The results are given in macro-$F_1$.}
\label{fig:nli-english-hs}
\end{figure*}

\subsection{NLI Fine-Tuning} 
In order to test the effectiveness of NLI fine-tuning, we add to this setup an additional optional phase which is placed before \textbf{En HS}.
This optional first phase has three variants: 
\begin{description}
\itemsep-0.2em 
\item[M + NLI:] A monolingual model in the target language is fine-tuned on the subset of XNLI \cite{conneau-etal-2018-xnli}\footnote{More information on the XNLI dataset is given in Appendix \ref{appsec:datasets}.} that is also in the target language. 
\item[X + NLI:] The multilingual model is fine-tuned on the subset of XNLI that is in the target language. 
\item[X + XNLI:] The multilingual model is fine-tuned on the entire XNLI dataset, concatenated with the MNLI dataset \cite{williams-etal-2018-broad}\footnote{More information on MNLI is given in Appendix \ref{appsec:datasets}.}. To encourage cross-lingual transfer learning, the translations of premises and hyptheses are shuffled such that for a given example the premise might be in Spanish and the hypothesis in Arabic.\footnote{This method of shuffling translations such that the premise and hypothesis are presented in different languages to the model has been employed for popular models such as \href{https://huggingface.co/joeddav/xlm-roberta-large-xnli}{joeddav/xlm-roberta-large-xnli}.}
\end{description}
If a model has been trained on NLI examples, we continue to train that model in an NLI formulation, even when fine-tuning on hate speech data. The model is then presented with the input text as the premise and ``This text is hate speech.'' as the hypothesis. The label ``hate speech'' then corresponds to ``entailment'' and ``not hate speech'' to ``contradiction''.

\subsection{Hypothesis Engineering}
\label{subsec:hypo-eng}
We employ the combination of the three strategies ``Filtering by Target'', ``Filtering Counterspeech'', and ``Filtering Reclaimed Slurs'' since they led to the best results in \citet{goldzycher-schneider-2022-hypothesis}. Further implementation details are provided in Appendix \ref{appsec:hypo-eng-details}. We evaluate the strategies in combination with all models that have been fine-tuned on XNLI or a subset of it. All models that were fine-tuned on a hate speech dataset are specific to one hypothesis claiming that there is hate speech in the input text. In such cases, we thus use the initial model, that has only been fine-tuned on an NLI dataset for all auxiliary hypothesis predictions.

\section{Evaluation}
\label{sec:evaluation}
\begin{figure*}[t]
\center
\includegraphics[width=\linewidth]{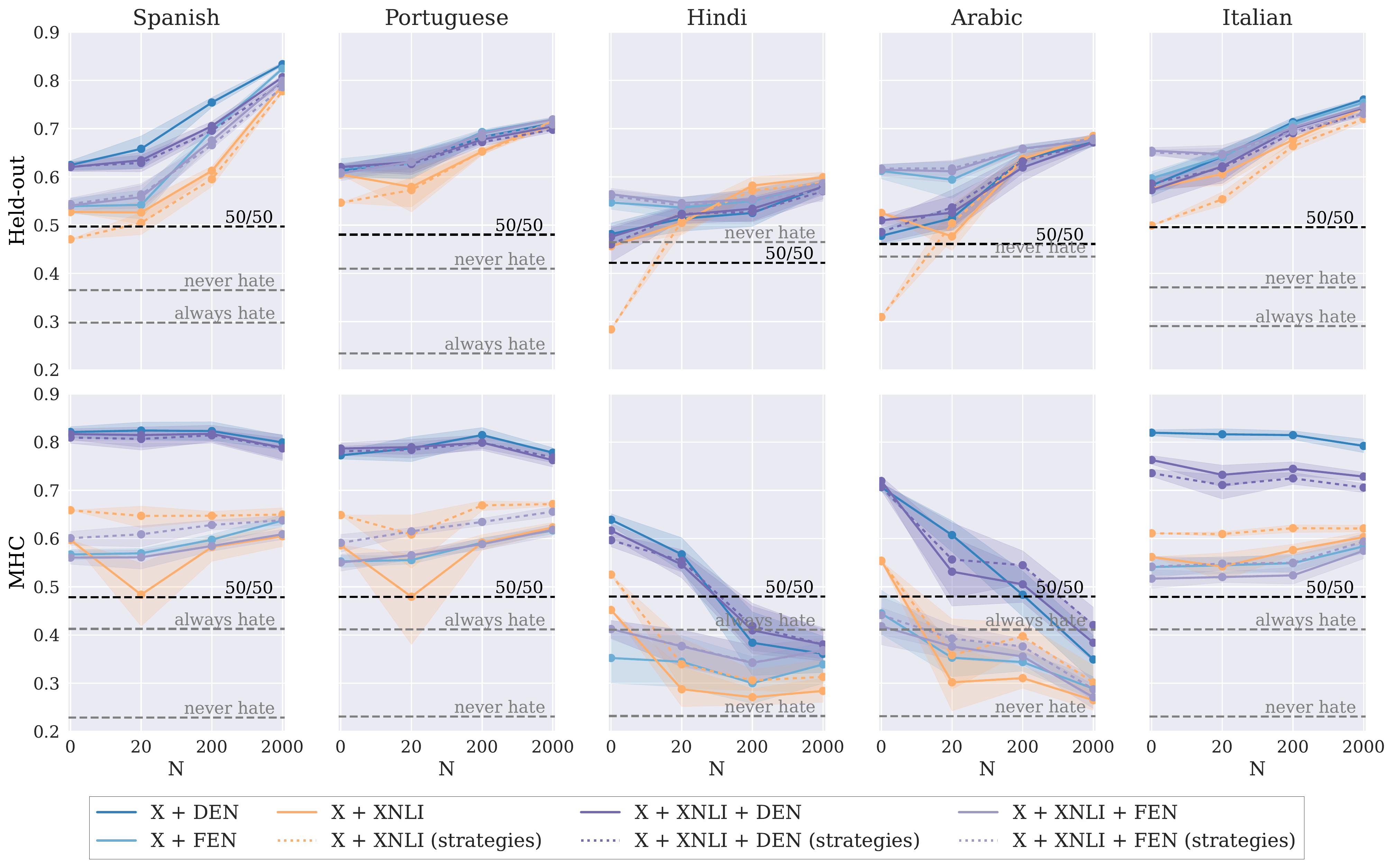}
\caption{Evaluation of hypothesis engineering strategies.}
\label{fig:results-strategies}
\end{figure*}

Following \citet{rottger-etal-2022-data}, we evaluate each setting displayed in Figure \ref{fig:experiment-setup} on two test-sets: (1) the held-out test set in the target language and (2) the HateCheck dataset in the target language. But in contrast to their setup, we evaluate in $N \in \{0, 20, 200, 2000\}$ target language training examples. We thus evaluate at three scenarios ($20$, $200$ and $2000$) where our results are directly comparable to the results of \citet{rottger-etal-2022-data} and add a zero-shot scenario.  
The metric is macro-$F_1$ in order to account for imbalanced test sets. Like \citet{rottger-etal-2022-data}, we train 10 models per setting, report the averaged results and the bootstrapped 95\% confidence intervals, represented as errorbars in Figure \ref{fig:baseline-nli-english-hs-seen-langauges} and shaded areas in Figures \ref{fig:nli-english-hs} to \ref{fig:full-results-mhc}. 

In what follows, we group the results according to research questions.
The full results are given in Appendix \ref{appsec:full-results}.

\subsection{Can we Reproduce the Results of \citet{rottger-etal-2022-data}?}
\label{subsec:reproduction-results}

Table \ref{tab:repro-results} contains the reproduction results and the average differences to the original results per language, number of examples $N$, and model. On average our results are lower by two percentage points on the held-out test sets and higher by one percentage point on the HateCheck test sets.  
We observe that: 
(1) Like \citet{rottger-etal-2022-data}, our results follow a trend of diminishing returns. The larger performance increase often comes from increasing from 20 to 200 examples and not from increasing from 200 to 2000 examples, even though the absolute increase in examples is much larger in the second comparison.
(2) Like \citet{rottger-etal-2022-data}, we see that with an increasing number of examples, the benefit of fine-tuning on English hate speech decreases. At 2000 examples, the monolingual model that has directly been fine-tuned on target language examples has in most cases caught up with or even beats English fine-tuning.

Overall, we view our results as a confirmation of their findings.
In order to simplify our evaluation, and since \textbf{X + FEN} and \textbf{X + KEN} have very similar results, we will only include \textbf{X + FEN} as a representative for natural data (as opposed to synthetic, adversarial data) in the following experiments.

\subsection{How Does an NLI-Formulation Compare to \citet{rottger-etal-2022-data}?}

We compare the baselines and settings proposed by \citet{rottger-etal-2022-data} with fine-tuning on monolingual NLI data in the target language (\textbf{M + NLI} and \textbf{X + NLI}) and fine-tuning on the entire XNLI dataset (\textbf{X + XNLI}). Only three languages (Arabic, Hindi, and Spanish) appear both in the evaluation setup and in XNLI. Since \textbf{M + NLI} and \textbf{X + NLI} training is only possible for languages in XNLI, we focus on the results in these three languages. The differences in performance averaged over the three languages are given in Figure \ref{fig:baseline-nli-english-hs-seen-langauges}.

Overall we see that introducing an intermediate fine-tuning step improves performance in most cases, but the benefits decrease with more target language examples available.
While \textbf{X} is almost always improved by NLI fine-tuning, the stronger (recall from Table \ref{tab:repro-results}) baseline \textbf{M} only benefits from NLI fine-tuning on the HateCheck testset (second row of plots).
When comparing \textbf{X + NLI} with \textbf{X + XNLI} we observe slightly more benefits from fine-tuning on the full XNLI dataset. 
Even though NLI-fine-tuning (orange) leads to clear benefits it is outperformed by fine-tuning on English hate speech on \textbf{X} (blue) as proposed by \citet{rottger-etal-2022-data} in all setups. This finding raises the question if training on additional English labeled data is also beneficial in an XNLI-based setup which we aim to answer in the next section.

\subsection{Are there Benefits to NLI-Finetuning when Given English Hate Speech Data?}

To answer this question we compare the performance of intermediate fine-tuning on English hate speech \citep{rottger-etal-2022-data} to first fine-tuning on XNLI and then fine-tuning on English hate speech. The results are displayed in Figure \ref{fig:nli-english-hs}.

We make the following observations:
(1) Since we now evaluate on all five languages, it is interesting to see how an NLI formulation performs for Portuguese and Italian which are not part of the XNLI dataset. For both testsets, we observe that even for unseen languages an NLI formulation has clear benefits over standard finetuning on target language examples (orange vs.\ green).
(2) On held-out test sets (first row of plots), fine-tuning on both XNLI and English hate speech (purple) improves zero-shot performance in Hindi, Arabic and Italian and matches zero-shot performance of only fine-tuning on English hate speech (blue) in Spanish and Portuguese. 
(3) The combination of the two approaches (purple) can also lead to improved few-shot results on Arabic and Hindi but reaches comparable performance or has a negative effect, when compared to only English hate speech fine-tuning (blue), in the other languages.
(4) On Multilingual HateCheck (second row of plots) we observe mixed results, where additional intermediate XNLI fine-tuning tends to decrease results when fine-tuning on \textbf{DEN} but increase results when fine-tuning on \textbf{FEN}.

Overall, we find that performance improves more from intermediate fine-tuning on English (green vs.\ blue) than an NLI formulation (green vs.\ orange) and that there are negligible advantages to combining the two approaches (blue vs.\ purple). As discussed in the related work section, previous work by \citet{goldzycher-schneider-2022-hypothesis} showed that zero-shot hate speech detection for English could be improved with carefully engineered auxiliary hypotheses for an NLI setup. In the next section, we focus on the question whether hypothesis engineering is also beneficial for our zero-shot and few-shot setups with other languages.

\subsection{Can Hypothesis Engineering further Improve Performance?}
We evaluate if hypothesis engineering, specifically the three strategies proposed by \citet{goldzycher-schneider-2022-hypothesis} as described in Section \ref{subsec:hypothesis-engineering}, is able to improve results. 
We take all models that have been fine-tuned on the full XNLI dataset and compare their performance with and without such hypothesis engineering strategies. The results are given in Figure \ref{fig:results-strategies}.

On the held-out test sets (first row of plots), the strategies (dotted lines) show only in a few cases small positive effects but mostly negative effects.
However, on HateCheck (second row of plots), we see that the strategies lead to clear performance improvements of \textbf{X + XNLI} and \textbf{X + XNLI + FEN}. This finding is not surprising since \citet{goldzycher-schneider-2022-hypothesis} specifically developed their hypothesis engineering strategies based on an error analysis on the English HateCheck data. For many languages, hypothesis engineering without intermediate English fine-tuning (orange dotted) even performs better than hypothesis engineering with fine-tuning on \textbf{FEN} (light purple dotted) which consists of Twitter data. Fine-tuning on English adversarial \textbf{DEN} hate speech examples (dark blue) remains the strongest approach for HateCheck since this fine-tuning data matches the testset conditions. However, our results show that if we have knowledge about the data the model will see at inference time (e.g.\ adversarial examples) but do not have matching English fine-tuning data (e.g.\ if only \textbf{FEN} were available),  hypothesis engineering geared towards the target domain (orange dotted line) is the best-performing approach.

\section{Conclusion}
\label{sec:conclusion}
In this work, we systematically evaluated the effectiveness of an NLI task formulation for hate speech detection in scenarios where only few labeled data in the target language are available. 
We were able to reproduce results by \citet{rottger-etal-2022-data}, who showed that intermediate fine-tuning on English hate speech is beneficial in such scenarios. 

Following their setup with our NLI-based experiments, we answered the following questions:
(1) How does NLI fine-tuning compare to English hate speech fine-tuning? 
Our results showed that while NLI fine-tuning leads to strong improvements over only fine-tuning in the target language it is outperformed by English hate speech fine-tuning.
(2) Are there benefits to combining NLI fine-tuning with English hate speech fine-tuning? 
We observed minor improvements when only $0$ or $20$ target language examples are available, but no benefits when more examples are available.
(3) Can hypothesis engineering further improve performance on the previously best NLI-based setting?
Our experiments demonstrated that hypothesis engineering can outperform other approaches only when the domain of the input data at inference time is known, but no matching training data is available.

Based on our results, we offer the following recommendations for hate speech detection in languages where little labeled data is available in the target language:

\begin{itemize}
    \item \textbf{In general domain scenarios:} Follow \citet{rottger-etal-2022-data} and perform standard intermediate fine-tuning on English data before training on target language examples if any are available.
    \item \textbf{In scenarios where less English data is available, e.g.\ hate speech against a specific protected group:} An intermediate NLI fine-tuning step is likely to be strongly beneficial compared to only fine-tuning on limited English and target language examples.
    \item \textbf{In scenarios where we have knowledge about the target domain but no matching English fine-tuning data is available:} Here, we suggest experimenting with targeted hypothesis engineering to reach the best possible performance. One exciting future avenue for this strategy is to focus on variation of protected groups across languages. Such culture and language-specific shifts will be hard to capture with English fine-tuning data but hypothesis engineering strategies offer more flexibility.
\end{itemize}
Finally, we want to highlight two areas for future work that arise from our experiments:
\begin{itemize}
    \item \textbf{Fine-tuning phases:} We believe that a crucial circumstance limiting the effectiveness of NLI models for hate speech detection is the fact that NLI training datasets are typically from different domains, thus creating a domain gap between NLI fine-tuning and other training phases. We hypothesize that by mixing the fine-tuning phases the negative effects of the domain gap might be avoided or, at least, reduced.
    \item \textbf{Hypothesis engineering and model capacity:} In contrast to \citet{goldzycher-schneider-2022-hypothesis}, we only observed positive results for hypothesis engineering in a few specific scenarios. One difference between the experiments in this paper and the ones in \citet{goldzycher-schneider-2022-hypothesis} that might explain the disparity in results is the model size: we use a base-sized multilingual RoBERTa model while they used an English-only \href{https://huggingface.co/facebook/bart-large-mnli}{BART-large model} \cite{lewis_bart_2020}, which we assume is generally more accurate in its NLI predictions and thus producing more reliable predictions for auxiliary hypotheses. Future work could thus evaluate the effect of model capacity on hypothesis engineering. 
\end{itemize}

\section*{Acknowledgments}
We thank Amit Moryossef and the anonymous reviewers for their helpful feedback.
Janis Goldzycher and Gerold Schneider were funded by the University of Zurich Research Priority Program (project ``URPP Digital Religion(s)''\footnote{\url{https://www.digitalreligions.uzh.ch/en.html}}). Chantal Amrhein received funding from the Swiss National Science Foundation (project MUTAMUR; no. 176727).

\section*{Limitations}
Even though our findings are backed by a large number of settings and experiments, the conclusions that can be drawn from this setup are limited in the following ways:

\paragraph{Datasets} We evaluated on five languages, a small fraction of the languages that could benefit from such models. Specifically, our results are limited to languages that appear in the pretraining dataset of XLM-T. Further, we showed that XNLI fine-tuning can lead to significant performance increases, even for languages that do not appear in XNLI. However, the two languages we tested on (Italian and Portuguese) are related to other languages in XNLI. There is a need for further research on how much of these benefits can be retained when the target language is less related to one of the languages in XNLI. 

\paragraph{Models} For consistency and efficiency, we used the same multilingual model in all experiments. This means that our results are dependent on the specifics of the model, specifically the domain it has been pre-trained on and its model size. As already mentioned in Section \ref{sec:conclusion}, the small model size likely has had a negative impact on the results for hypothesis engineering. 


\paragraph{Hypothesis Engineering} All results are limited to the specific strategies that we tested. Further, potential errors in the automatic translations of the hypotheses of the strategies might have impacted the results. A strength of the approach, that we did not evaluate, is that they can be adjusted to new languages and cultures simply by specifying e.g. a new set of protected groups or group characteristics via auxiliary hypotheses, thereby avoiding the criticism that zero-shot cross-lingual hate speech detection does not adjust to the specific circumstances of a language and culture \cite{nozza-2022-nozza}.

\newpage

\bibliography{custom}
\bibliographystyle{acl_natbib}

\appendix

\newpage

\section{Training Details}
\label{appsec:training-details}

All models were trained and evaluated using the \href{https://huggingface.co/docs/transformers/index}{\texttt{transformers}} library \cite{wolf-etal-2020-transformers}, version \texttt{4.26.1}. 
The hyperparameters for the NLI-fine-tuning stage are provided in Table \ref{tab:nli-fine-tuning}.
\begin{table}[h]
\centering
\begin{tabular}{lr}
\toprule
parameter & \multicolumn{1}{l}{value} \\
\midrule
epochs & 5 \\
learning rate & 2e-05 \\
batch size & 32 \\
max sequence length & 128 \\
\bottomrule
\end{tabular}
\caption{Hyperparameters of NLI fine-tuning. }
\label{tab:nli-fine-tuning}
\end{table}
We always chose the best performing checkpoint on the validation set out the five epochs.

For intermediate fine-tuning on English hate speech and fine-tuning on target language examples, we used the same hyperparameters as \citet{rottger-etal-2022-data}. They are given in Table \ref{tab:en-hs-fine-tuning} and Table \ref{tab:tl-hs-fine-tuning}.

\begin{table}[h]
\centering
\begin{tabular}{lr}
\toprule
parameter & \multicolumn{1}{l}{value} \\
\midrule
epochs & 3 \\
learning rate & 5e-05 \\
batch size & 16 \\
max sequence length & 128 \\
\bottomrule
\end{tabular}
\caption{Hyperparameters for fine-tuning on English hate speech datasets. }
\label{tab:en-hs-fine-tuning}
\end{table}

\begin{table}[h]
\centering
\begin{tabular}{lr}
\toprule
parameter & \multicolumn{1}{l}{value} \\
\midrule
epochs & 5 \\
learning rate & 5e-05 \\
batch size & 16 \\
max sequence length & 128 \\
\bottomrule
\end{tabular}
\caption{Hyperparameters for fine-tuning on hate speech datasets in the target language. }
\label{tab:tl-hs-fine-tuning}
\end{table}

All other hyperparameters were kept at the default values of the huggingface Trainer-class.

\section{Hypothesis Engineering Details}
\label{appsec:hypo-eng-details}

Since we worked with monolingual models in the target language and with multilingual models that have been fine-tuned on English and other languages first, this raises the question in which language the hypotheses should be expressed.
For monolingual models, we automatically translated all hypotheses with Google Translate to the target language. The model thus received the premise and hypothesis in the same language as input.
For multilingual models, we kept the original English hypotheses. Since we shuffled the languages of premise and hypothesis in the NLI training-regime the models should be able to handle the differing languages well. Keeping the hypotheses in multilingual models in English also means that the hypothesis remains in the same language over multiple fine-tuning phases like intermediate fine-tuning on English hate speech data and target language fine-tuning.

\citet{goldzycher-schneider-2022-hypothesis} proposed two versions of the strategy ``Filtering by Target'': In the first version the model predicts if protected groups are targeted (e.g. ``This text is about Muslims.''). In the second version the model predicts if protected group characteristics are targeted (e.g. ``This text is about religion.''). Even though the second version performed worse in their experiments, we use this second version for our experiments, because its predictions are more neutral with respect to specific languages and cultures. This enabled us to use exactly the same strategies for each language. In a more sophisticated setup one could implement the first version predicting protected groups and adjust these groups for each language.

\section{Datasets}
\label{appsec:datasets}
The key characteristics of all datasets we used in the experiments are described in Table \ref{tab:datasets}. 
To create the MultiNLI dataset, sentences from diverse genres were collected and used as premises. Annotators then were tasked with creating artificial hypotheses for these premises. 
For XNLI, the test set was translated by human translators and the  training set was translated automatically. 

\begin{table*}[t]
\setlength\tabcolsep{3pt}
\centering
\resizebox{0.9\textwidth}{!}{
\begin{tabular}{llrrrrl}
\toprule
\multicolumn{1}{c}{code} & \multicolumn{1}{c}{paper} & \multicolumn{1}{c}{train} & \multicolumn{1}{c}{validation} & \multicolumn{1}{c}{test} & \multicolumn{1}{c}{\% hate} & \multicolumn{1}{c}{source}  \\ 
\midrule
\multicolumn{7}{c}{(X)NLI datasets} \\
MNLI & \text{\citet{williams-etal-2018-broad}} & 40000 & 19650 & \multicolumn{1}{c}{-} & \multicolumn{1}{c}{-} & diverse \\
XNLI & \text{\citet{conneau-etal-2018-xnli}} & 393000 & 24900 & \multicolumn{1}{c}{-} & \multicolumn{1}{c}{-} & translation of MNLI\\ 
\hdashline
\multicolumn{7}{c}{English hate speech datasets (En HS)} \\
FEN & \text{\citet{founta_large_2018}} & 20068 & 500 & \multicolumn{1}{c}{-} & 22.0 & Twitter \\ 
KEN & \text{\citet{kennedy-etal-2020-contextualizing}} & 20692 & 500 & \multicolumn{1}{c}{-} & 50.0 & Youtube, Twitter, Reddit \\  
DEN & \text{\citet{vidgen-etal-2021-learning}} & 38644 & 500 & \multicolumn{1}{c}{-} & 53.9 & annotators, adversarial \\ 
\hdashline
\multicolumn{7}{c}{Target Language Hate Speech Datasets (TL HS)} \\
BAS19\_ES & \text{\citet{basile-etal-2019-semeval}} & 4100 & 500 & 2000 & 41.5 & Twitter \\ 
FOR19\_PT & \text{\citet{fortuna-etal-2019-hierarchically}} & 3170 & 500 & 2000 & 31.5 & Twitter \\
HAS21\_HI & \text{\citet{10.1145/3503162.3503176}} & 3794 & 300 & 500 & 12.3 & Twitter\\
OUS19\_AR & \text{\citet{ousidhoum-etal-2019-multilingual}}  & 2053 & 300 & 1000 & 22.5 & Twitter\\
SAN20\_IT & \text{\citet{manuela2020haspeede}} & 5600 & 500 & 2000 & 41.8 & Twitter\\
\hdashline
\multicolumn{7}{c}{Multilingual HateCheck (MHC)} \\
HateCheck\_ES & \text{\citet{rottger-etal-2022-multilingual}} & \multicolumn{1}{c}{-} & \multicolumn{1}{c}{-} & 3745 &  70.3 & annotators \\
HateCheck\_PT & \text{\citet{rottger-etal-2022-multilingual}} & \multicolumn{1}{c}{-} & \multicolumn{1}{c}{-} & 3691 & 69.9 & annotators\\ 
HateCheck\_HI & \text{\citet{rottger-etal-2022-multilingual}} & \multicolumn{1}{c}{-} & \multicolumn{1}{c}{-} & 3565 & 69.8 & annotators \\ 
HateCheck\_AR & \text{\citet{rottger-etal-2022-multilingual}} & \multicolumn{1}{c}{-} & \multicolumn{1}{c}{-} & 3570 & 69.9 & annotators\\ 
HateCheck\_IT & \text{\citet{rottger-etal-2022-multilingual}} & \multicolumn{1}{c}{-} & \multicolumn{1}{c}{-} & 3690 & 70.0 & annotators \\
\bottomrule
\end{tabular}
}
\caption{Statistics of training and evaluation datasets. 
In the case of FEN and KEN, we applied downsampling to the non-hate speech class. The table reflects the state after downsampling.
}
\label{tab:datasets}
\end{table*}

\section{Full Results}
\label{appsec:full-results}

\begin{figure*}[t]
\center
\includegraphics[width=\linewidth]{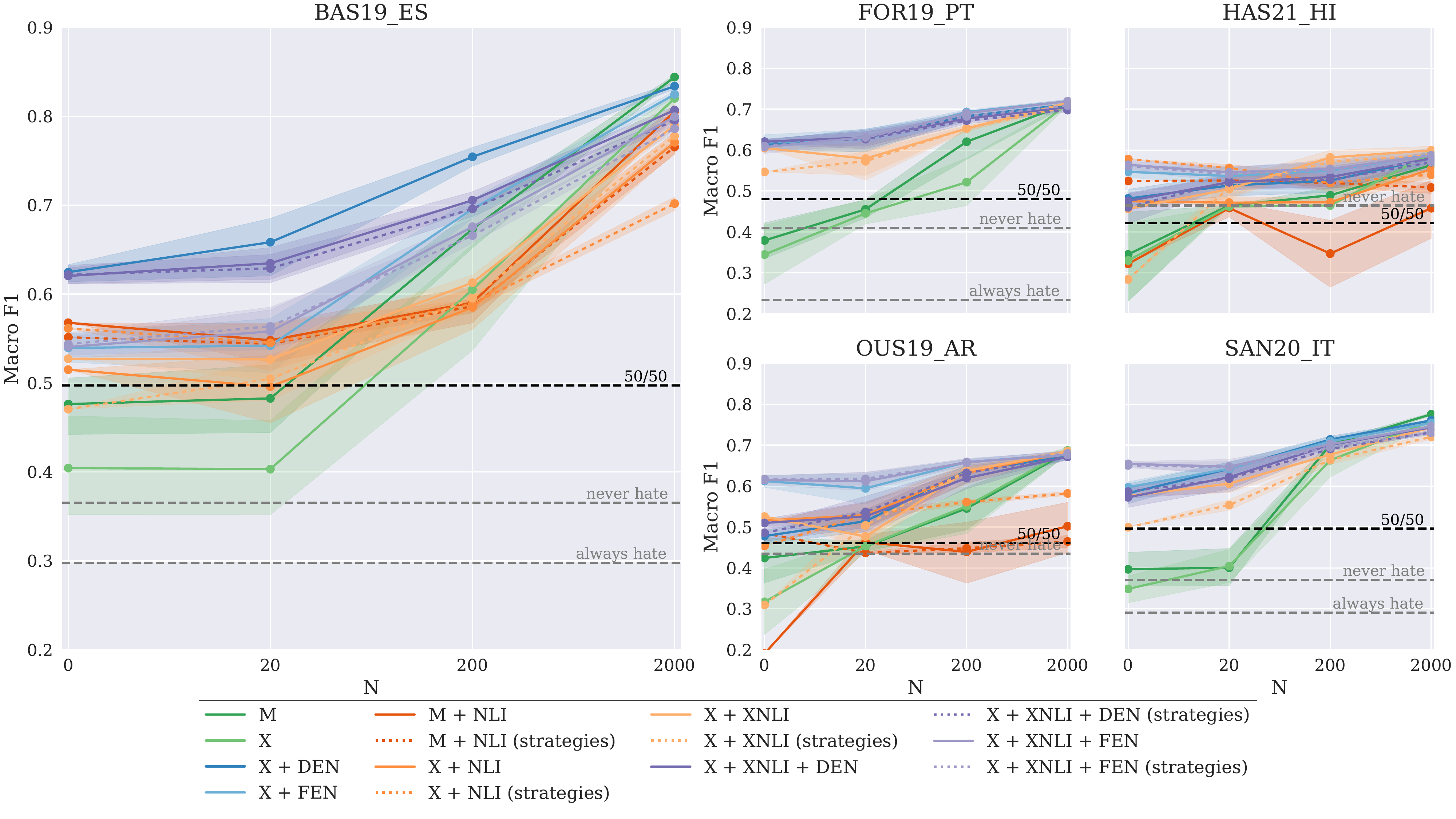}
\caption{The full results on held-out test sets.}
\label{fig:full-results-held-out}
\end{figure*}

\begin{figure*}[t]
\center
\includegraphics[width=\linewidth]{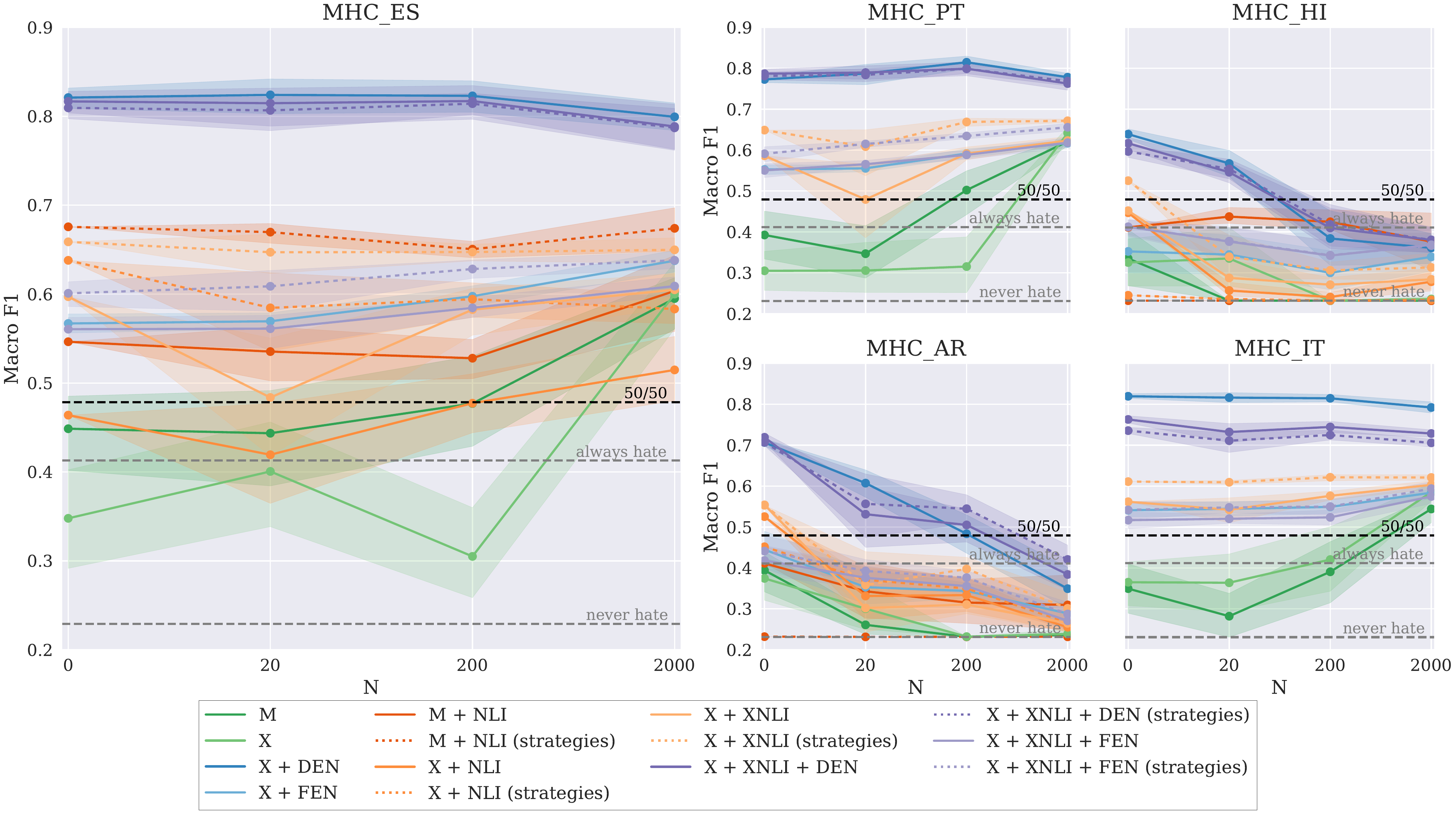}
\caption{The full results on the multilingual HateCheck.}
\label{fig:full-results-mhc}
\end{figure*}

In order to highlight specific aspects of our results, we split them up over several Figures in the paper. The full results of all settings that we evaluated are provided in Figures \ref{fig:full-results-held-out} and \ref{fig:full-results-mhc}. 

\end{document}